\documentclass[conference]{IEEEtran}
\IEEEoverridecommandlockouts
\usepackage{cite}
\usepackage{amsmath,amssymb,amsfonts}
\usepackage{algorithmic}
\usepackage{graphicx}
\usepackage{textcomp}
\usepackage{xcolor}
\usepackage{enumerate}
\usepackage{array,booktabs,arydshln,xcolor}
\usepackage[hidelinks]{hyperref}
\usepackage{caption}
\usepackage{enumitem}
\usepackage{tablefootnote}
\usepackage{comment}

\newcommand{\cmmnt}[1]{}

\def\BibTeX{{\rm B\kern-.05em{\sc i\kern-.025em b}\kern-.08em
    T\kern-.1667em\lower.7ex\hbox{E}\kern-.125emX}}

\begin{document}

\title{Smart Sensing for Safer Bridges: From Sensor Signals to AI-Driven Anomaly Detection \\
}

\author{
    \IEEEauthorblockN{Rahul Jaiswal, Joakim Hellum,  Halvor Heiberg}
    \IEEEauthorblockA{
    Smart Sensor Systems AS \\
    Oslo, Norway \\
    \{rahul.jaiswal, joakim.hellum, halvor.heiberg\}@smartsensorsystems.no}
}

\IEEEoverridecommandlockouts \IEEEpubid{\makebox[\columnwidth]{979-8-3315-8949-3/26/\$31.00 \copyright 2026 IEEE \hfill} \hspace{\columnsep}\makebox[\columnwidth]{ }}

\maketitle

\begin{abstract} 
Bridges contribute significantly to transportation connectivity and urban development. Therefore, reliable bridge monitoring is crucial for protecting public safety and detecting anomalous behavior in bridge sensor data that may provide early indications of abnormal structural conditions. This paper investigates anomaly detection in real-world bridge sensor data using two different complementary approaches, namely signal processing and the data-driven machine learning model Isolation Forest. The real-time bridge sensor data is collected from an iBridge sensor device installed on a bridge in Norway. The methods are evaluated using anomaly counts, anomaly detection time, processing rate, anomaly rates, visualization, and temporal agreement. Moreover, a controlled anomaly-injection analysis is performed to evaluate the sensitivity of each method. Numerical results demonstrate distinct detection characteristics and computational requirements, highlighting the potential of machine learning, particularly the data-driven Isolation Forest, alongside signal processing for identifying anomalies in bridge sensor measurements.
\end{abstract}

\section{Introduction} 
Bridges constitute an essential element of smart-city infrastructure and play a vital role in supporting modern transportation networks. With hundreds of thousands of bridges in operation worldwide, maintaining their structural safety and serviceability is of critical importance. Conventional bridge condition assessment has predominantly depended on manual visual inspections. However, these inspections can be labour-intensive and susceptible to human error, particularly when identifying and quantifying deterioration in concrete and wooden bridge structures. To address these limitations, intelligent sensor-based and data-driven machine learning approaches have increasingly been adopted for bridge monitoring. Such techniques can facilitate timely evaluation of structural conditions, improve public and pedestrian safety, and reduce the likelihood of severe structural failures and related accidents. Consequently, reliable bridge monitoring requires intelligent models which are capable of distinguishing normal structural behaviour from anomalous sensor data responses.

\begin{figure} [t!]
  \centering
  \includegraphics[width=0.95\columnwidth, height=5.0cm, keepaspectratio]{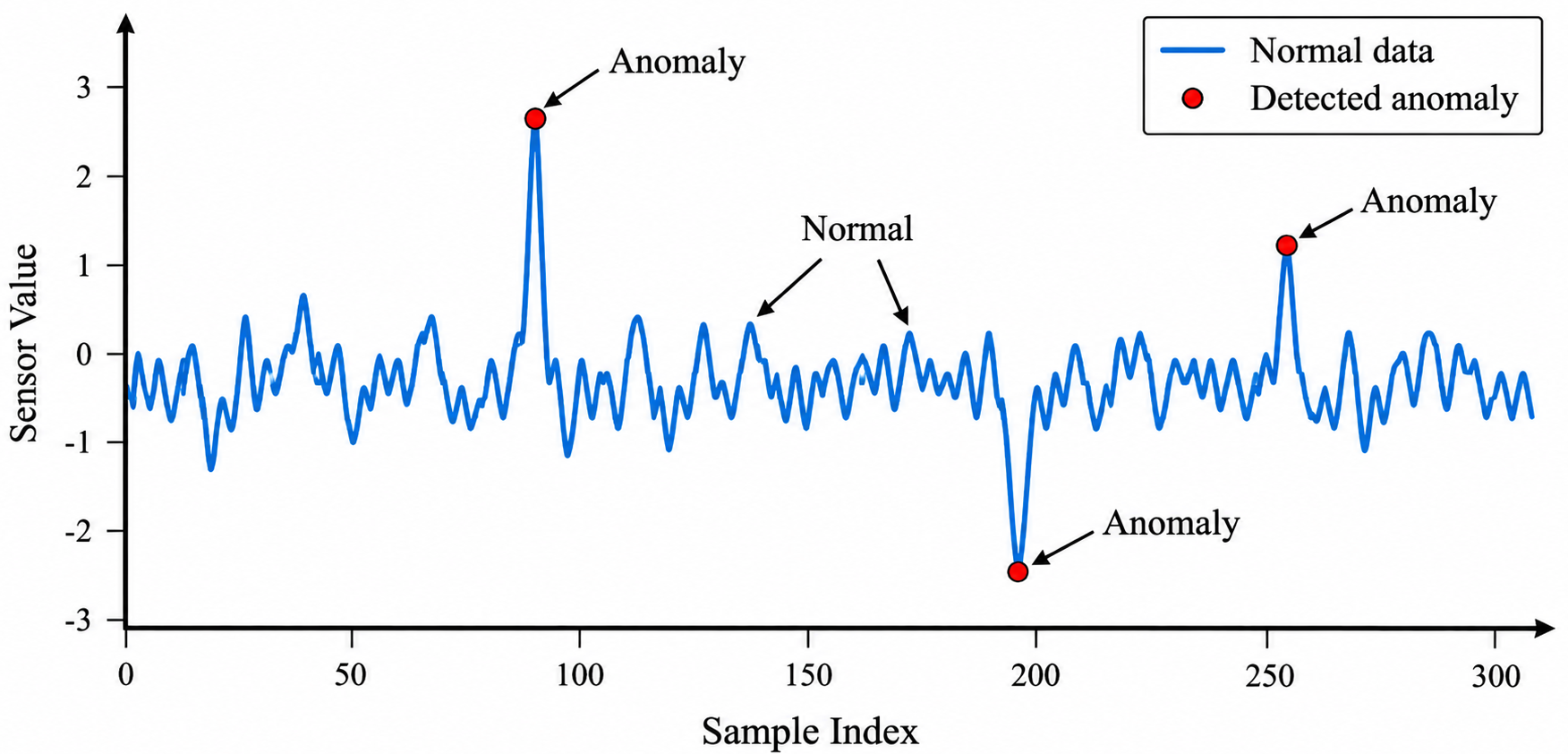}
  \caption{Example of anomaly detection in sensor data.}
  \label{anomaly_exm}  \vspace{-4mm}
\end{figure}

Anomaly detection is a key data-mining task~\cite{han2022data}, commonly referred to as outlier detection. For bridge monitoring systems, an anomaly represents a sensor observation that departs noticeably from the established pattern of normal measurements. Such deviations can be associated with structural damage, unusual loading, environmental influences, or sensor-related faults. The primary purpose of anomaly detection is to recognize these irregular observations and distinguish them from the expected behaviour of the bridge. Figure~\ref{anomaly_exm} shows a simple example of anomaly detection in sensor measurements, where the anomalous observation is characterized by a substantially larger peak than the surrounding normal measurements.

Various approaches have been explored for identifying anomalies in sensor measurements. Statistical approaches~\cite{samariya2023comprehensive}, including multivariate models~\cite{edirisinghe2015markov} and mean-variance-based methods~\cite{elton2009modern}, generally establish patterns of normal structural behaviour from historical monitoring data. Measurements that substantially depart from these established patterns are subsequently considered anomalous. However, the performance of such approaches often depends on prior knowledge of the data characteristics, including assumptions regarding the underlying distribution and appropriate model parameters. Since real-world sensor measurements may follow unknown or complex distributions, these assumptions can reduce the reliability of statistical-based anomaly detection in practical bridge monitoring scenarios.

To monitor bridges, different anomaly detection methods, including signal processing techniques, are used~\cite{sonbul2025bridge}. Artificial intelligence (AI), particularly machine learning (ML), has become increasingly prominent in detecting anomalies~\cite {jaiswal2025leveraging}. These methods learn informative patterns directly from raw sensor measurements, reducing the need for manually defined features. Neural network-based approaches are also used due to their powerful representation-learning and universal approximation capabilities~\cite{jaiswal2023caqoe}. In~\cite{hamedani2025bridge}, bridge monitoring uses unsupervised anomaly detection to identify structural damage. A convolutional neural network is applied to detect structural damage in bridge systems~\cite{abdeljaber2017real}, while an autoencoder-based approach is investigated for bridge monitoring in~\cite{tian2025anomaly}.

Bridge anomaly detection can preserve structural integrity, protecting public and pedestrian safety, and reducing the likelihood of severe failures or accidents. Based on these observations, this study investigates signal processing and machine learning approaches for anomaly detection in bridge sensor data. The proposed framework analyzes sensor measurements collected using an iBridge device installed on a bridge in Norway. Early identification of abnormal sensor patterns can facilitate timely maintenance actions and help reduce associated structural risks.

The paper is structured as follows. Section~\ref{ml_bgd} introduces the anomaly detection techniques considered in this study. Section~\ref{prop_model} details the proposed detection framework, while Section~\ref{dset} provides an overview of the dataset. Section~\ref {res_dis} presents the findings and their analysis. Finally, Section~\ref{cons} presents conclusions and future work.

\section{Background} 
\label{ml_bgd}
This section presents the techniques employed for anomaly detection in the bridge monitoring system.

\subsection{Signal Processing} 
\label{sig_algo}
\begin{figure} [t!]
  \centering
  \includegraphics[width=\columnwidth, height=5.5cm, keepaspectratio]{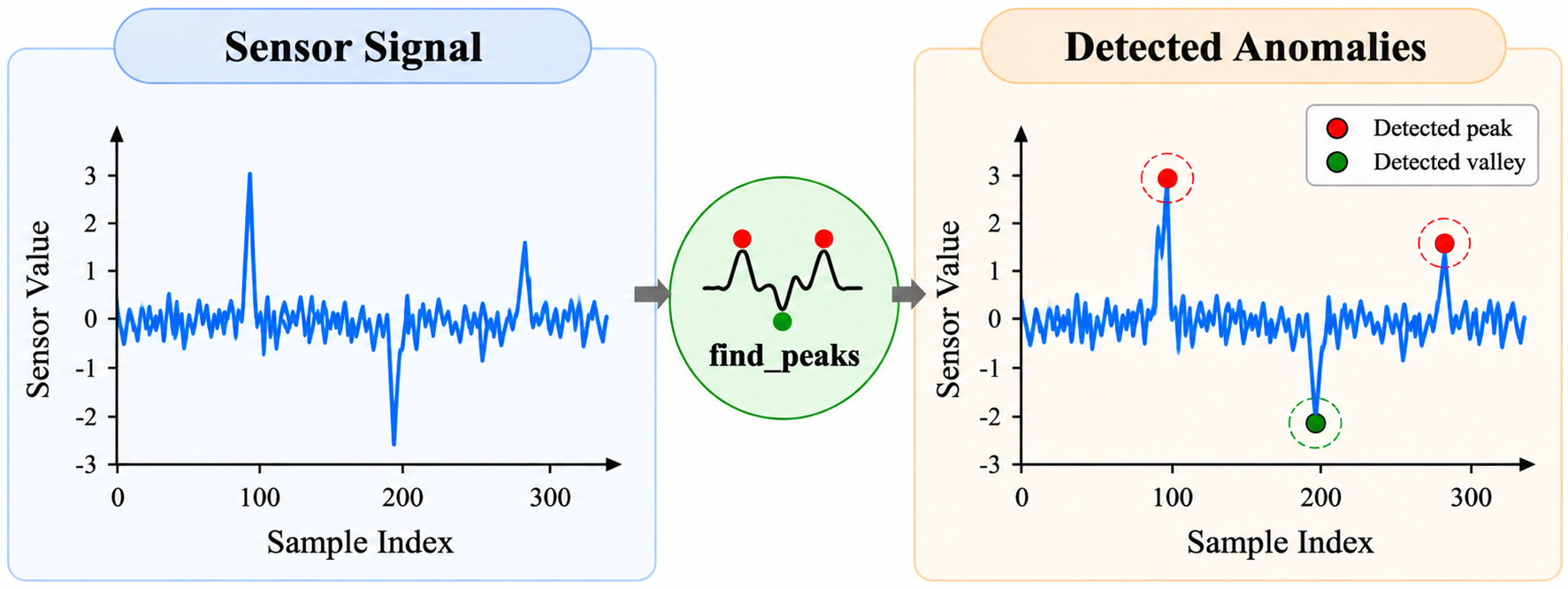}
  \caption{A simple illustration of the signal processing technique.}
  \label{sig_pro} \vspace{-3mm}
\end{figure}

The signal processing technique detects abnormal events by identifying significant peaks and valleys in the bridge sensor signals using the \texttt{find\_peaks} algorithm~\cite{sonbul2025bridge}. The algorithm identifies local maxima and minima that exhibit sufficient amplitude, prominence, and separation from neighbouring events. Parameters such as height, prominence, and minimum distance distinguish significant signal deviations from normal fluctuations. Negative valleys are detected by applying the method to the inverted signal. The detected peaks and valleys, along with their corresponding sample indices, are then recorded as potential anomalies, as shown in Fig.~\ref{sig_pro}.

\subsection{Isolation Forest} 
\label{if_algo}
\begin{figure} [t!]
  \centering
  \includegraphics[width=\columnwidth, height=5.5cm, keepaspectratio]{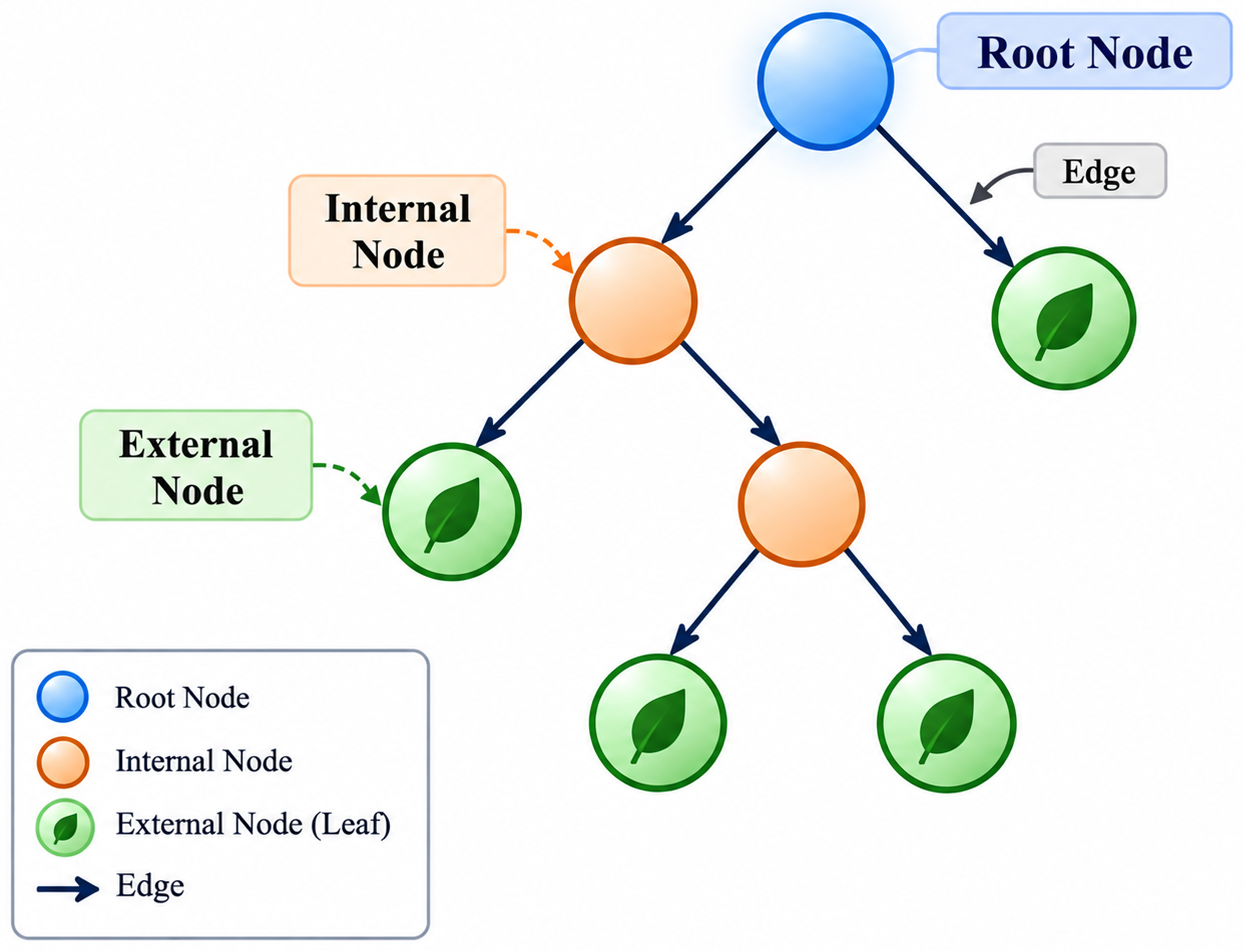}
  \caption{Illustration of a typical Isolation Tree (iTree) structure.}
  \label{IF_tree} \vspace{-3mm}
\end{figure}

The Isolation Forest (IF) algorithm is a data-driven machine learning model that identifies unusual observations (anomalies) by combining the outputs of multiple randomly constructed binary trees~\cite{chabchoub2022depth}. For bridge sensor data, each tree recursively partitions the sensor observations by randomly selecting a feature and a corresponding split point. A collection of such isolation trees (iTrees) forms the overall isolation forest~\cite{xu2023deep}. Through successive random partitioning, observations are separated based on how easily they can be isolated from the remaining data. Anomalous observations generally require fewer partitioning steps and therefore appear closer to the root of an iTree, while observations representing normal operating conditions typically require longer paths and occur deeper within the tree. A simplified example of an iTree~\cite{jaiswal2025leveraging} is presented in Fig.~\ref{IF_tree}. In this iTree, an external node represents a terminal node without children, whereas an internal node branches into two child nodes. Consequently, observations reaching external nodes after relatively short paths are typically regarded as anomalous.

\section{Proposed Model} 
\label{prop_model}
\begin{figure*} [t!]
  \centering
  \includegraphics[width=0.85\textwidth, height=9.5cm]{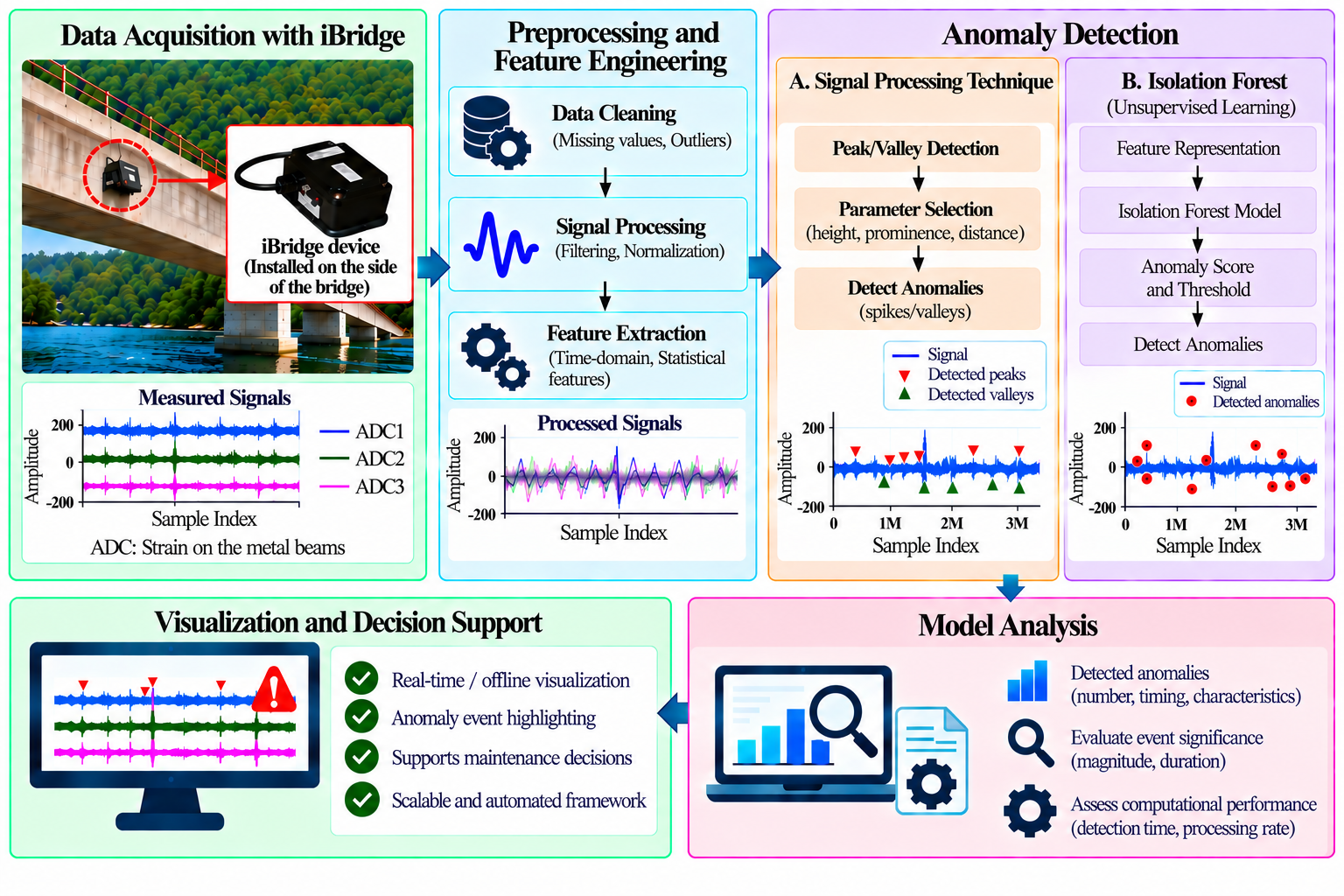}
  \caption{Overview of the proposed bridge anomaly detection model.}
  \label{prop_mod} \vspace{-3mm}
\end{figure*}

The proposed framework, illustrated in Fig.~\ref{prop_mod}, provides a systematic pipeline for detecting abnormal behaviour in bridge sensor data collected using the iBridge sensor device (see Section~\ref{data_coll}). The acquired sensor signals first undergo data cleaning and signal preprocessing, followed by feature extraction to obtain informative representations of the signals. Anomaly detection is then performed using a signal processing technique and an Isolation Forest model. The detected anomalies from both approaches are subsequently analyzed in terms of their number, timing, magnitude, duration, and computational performance, including anomaly detection time and processing rate. Finally, the identified anomalies are visualized to facilitate interpretation of abnormal sensor behaviour and provide information that can support bridge maintenance and monitoring decisions.

\section{Dataset} 
\label{dset}
This section presents the sensor device and dataset used for real-time bridge monitoring.

\subsection{iBridge Sensor Data Acquisition Device}
\label{data_coll}
\begin{figure} [t!]
  \centering
  \includegraphics[width=\columnwidth, height=4.5cm, keepaspectratio]{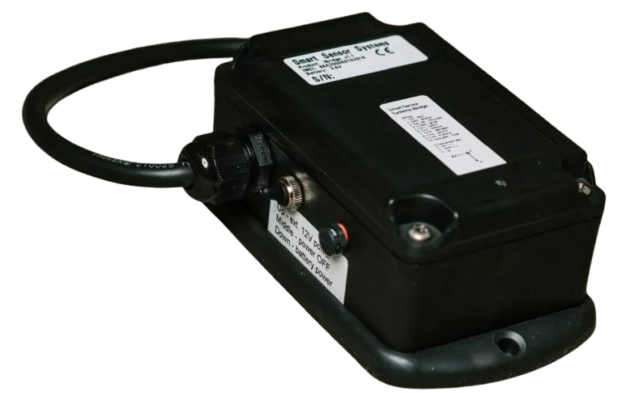}
  \caption{Illustration of the iBridge sensor device.}
  \label{ibridge_device} \vspace{-3mm}
\end{figure}

The bridge sensor data are acquired using iBridge, a sensor-based device illustrated in Fig.~\ref{ibridge_device}. The iBridge~\cite{ibridge_sss} is a compact monitoring unit that integrates multiple sensor interfaces and communicates the collected sensor measurements to a cloud platform through 4G connectivity. The device supports battery-powered operation and can be connected to various sensors and transmitters. It is primarily designed for strain-gauge applications and enables the measurement of a wide range of structural parameters, including acceleration, tensile and compressive stress, cracking, displacement, collapse, strain in metal beams, weight, pressure, vibration, angular variation, torque, temperature, and liquid level.

The iBridge device has built-in accelerometer and temperature sensors, enabling it to operate as a multifunctional data-logging unit. Its compact design allows convenient installation on different materials and surfaces, while its portable configuration supports relocation and repeated use. The device also offers flexible battery options to accommodate different deployment requirements. The direct sensor interfacing eliminates the need for external sensor cabling and fixed power connections. Furthermore, the iBridge integrates with an intelligent cloud platform for device configuration and data storage, enabling continuous 24/7 access to sensor data and facilitating integration with an enterprise resource planning (ERP) system.

\subsection{Bridge Sensor Data}
\label{bridge_data}
The bridge sensor data is acquired from a bridge in Inland, Norway. As illustrated in Fig.~\ref{prop_mod}, the iBridge sensor device is installed on the side of the bridge. The real-time measurements are collected over 11 days, that is, from June 10 to June 20, 2026. The dataset comprises downward strain measurements from the metal beams at three different locations, denoted as ADC1, ADC2 and ADC3, respectively, and recorded in mV. In total, the dataset contains 3,774,011 numerical samples, collected at a sampling rate of 5 Hz, corresponding to one measurement every 0.2 seconds.

\section{Results and Discussions} 
\label{res_dis}
This section describes the experimental setup and presents a detailed analysis of the obtained results.

\subsection{Experimental Setup} 
\label{sys_set}
All experiments are conducted in Python~3.13.6, using the Keras framework with TensorFlow~2.20.0. NumPy~2.3.2, Pandas~2.3.1, and scikit-learn~1.7.2 are employed for data preprocessing. We used a MacBook equipped with an Apple M4 processor and 16~GB of RAM.

\subsection{Data Preprocessing}
\label{data_prepro}
\begin{figure} [t!]
    \centering
    \includegraphics[width=\columnwidth, height=3.0cm]{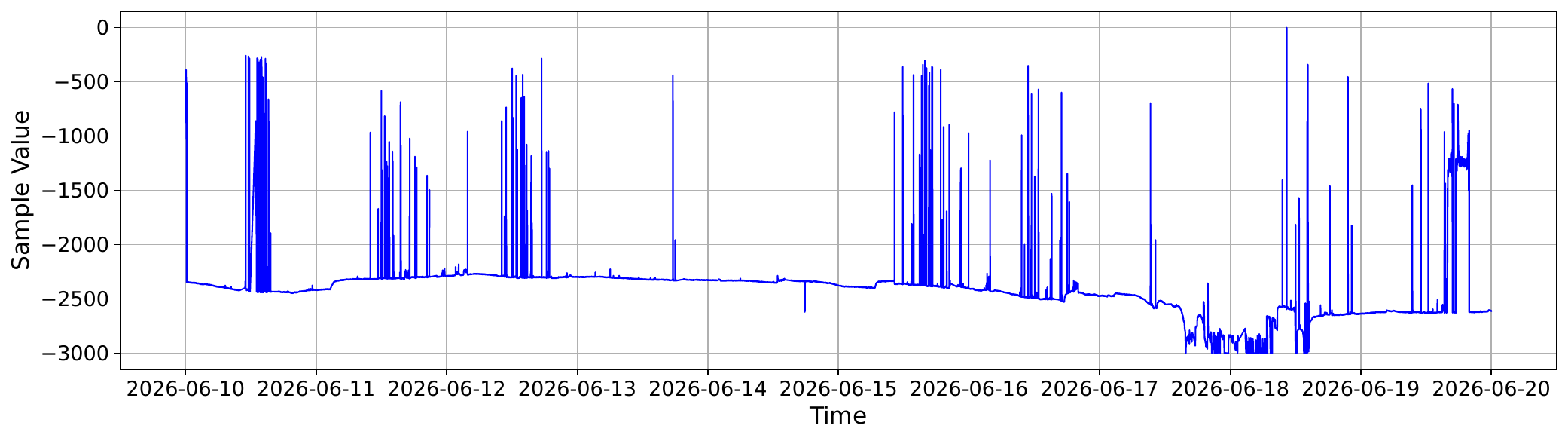}
    \caption{ADC1 signal in bridge sensor data.}
    \label{adc1_data} \vspace{-3mm}
\end{figure}

\begin{figure} [t!]
    \centering
    \includegraphics[width=\columnwidth, height=3.0cm]{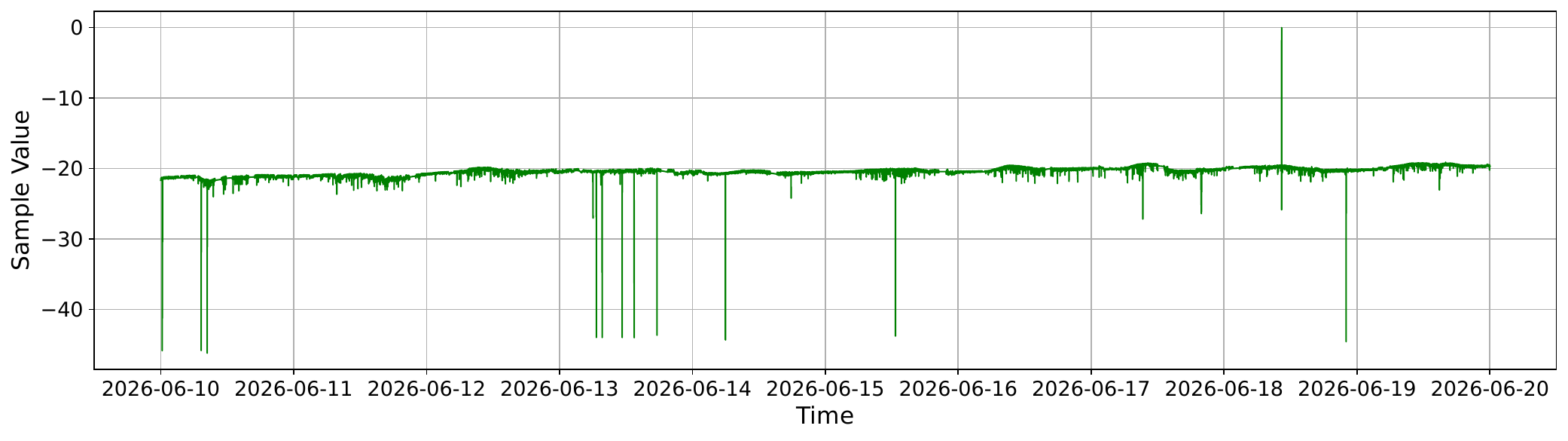}
    \caption{ADC2 signal in bridge sensor data.}
    \label{adc2_data} \vspace{-3mm}
\end{figure}

\begin{figure} [t!]
    \centering
    \includegraphics[width=\columnwidth, height=3.0cm]{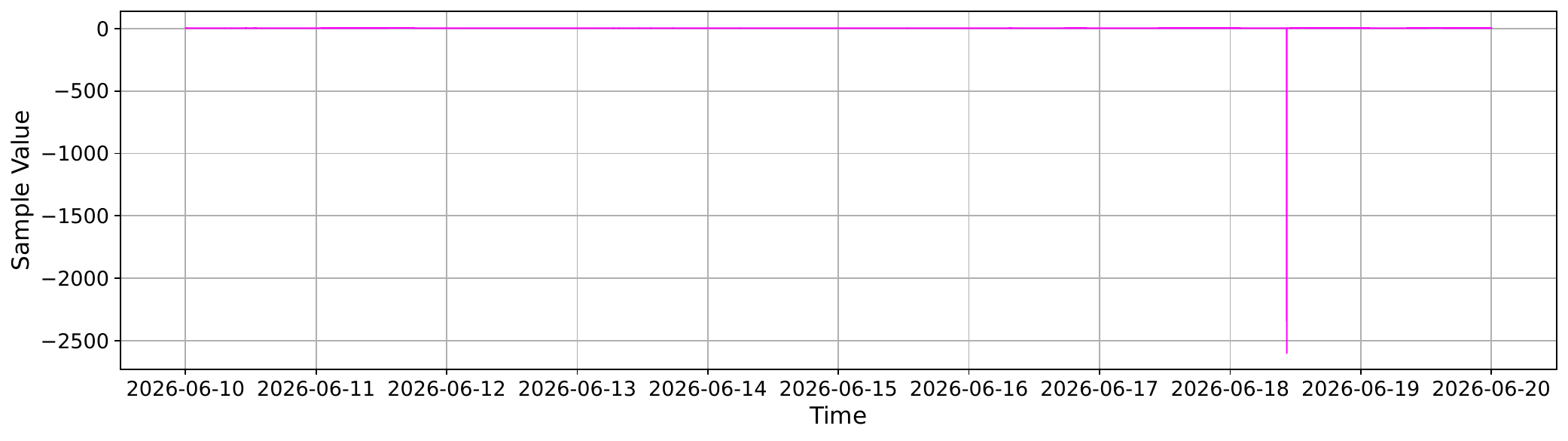}
    \caption{ADC3 signal in bridge sensor data.}
    \label{adc3_data} \vspace{-3mm}
\end{figure}

The preprocessing stage includes data cleaning and handling missing observations. Since incomplete sensor measurements can affect the reliability of subsequent model analysis, the collected bridge sensor data are processed to improve their quality and consistency. Only a small number of samples are missing in the dataset, with 285 entries in ADC1, 285 in ADC2, and 290 in ADC3. The missing observations are estimated using time-based linear interpolation~\cite{phillips2003interpolation}. Alternatively, one can also use transfer learning~\cite{jaiswal2023leveraging, jaiswal2023location,jaiswal2025data} approaches to estimate missing observations when appropriate. The final bridge signals are shown in Fig.~\ref{adc1_data}, Fig.~\ref{adc2_data}, and Fig.~\ref{adc3_data}, respectively. 

\subsection{Performance Evaluation} 
\label{eva_met} 
To evaluate the performance of each method, four metrics are considered: the number of detected anomalies, anomaly detection time (seconds), processing rate (samples/second), and anomaly rate (\%). The anomaly detection time reflects the computational time required by each method, while processing rate indicates the number of samples processed per second. The anomaly rate represents the proportion of observations identified as anomalous. Moreover, for qualitative assessment, the detected anomalies are also visualized in the corresponding plots. Since the bridge sensor data does not contain ground-truth anomaly labels, classification metrics such as precision, recall, and F1-score cannot be computed. The metrics are defined as follows:
\begin{equation}
T_{\mathrm{det}} = t_{\mathrm{end}} - t_{\mathrm{start}}
\end{equation}
where $T_{\mathrm{det}}$ denotes the total anomaly detection time. $t_{\mathrm{start}}$ and $t_{\mathrm{end}}$ are the times at which the detection process starts and finishes, respectively.

\begin{equation}
R_{\mathrm{proc}} = \frac{N}{T_{\mathrm{det}}}
\end{equation}
where $R_{\mathrm{proc}}$ denotes the processing rate, 
$N$ is the total number of input samples, and $T_{\mathrm{det}}$ is the anomaly detection time.

\begin{equation}
R_{\mathrm{anom}} =
\frac{N_{\mathrm{anom}}}{N} \times 100
\end{equation}
where $R_{\mathrm{anom}}$ denotes the anomaly rate, 
$N_{\mathrm{anom}}$ is the number of detected anomalous samples, and $N$ is the total number of input samples.

\subsection{Proposed Model Analysis} 
\label{ana_result} 
The bridge sensor data introduced in Section~\ref{bridge_data} is used to develop and evaluate both anomaly detection methods described in Section~\ref{ml_bgd}. Before model development, the data undergoes the preprocessing steps outlined in Section~\ref{data_prepro} to enhance data quality. A grid search~\cite{elgeldawi2021hyperparameter} is then used to determine suitable parameter configurations for each method. The resulting parameter settings are summarized in Table~\ref{para_used}.

\begin{table} [t!]
\caption{Parameter settings for each method.}
\label{para_used} 
\centering
{
\setlength\tabcolsep{1.0pt}
\begin{tabular}{|c|c|c|}
\hline
\multicolumn{3}{|c|}{Method with parameters} \\
\hline
Signal & Signal Processing (SP) & Isolation Forest (IF)\\
\hline
ADC1 & prominence=80, distance=20 & $\text{n\_estimators}$\tablefootnote{In Isolation Forest, $\text{n\_estimators}$ denotes the number of isolation trees, contamination specifies the expected proportion of anomalous samples, and $\text{n\_jobs}$ determines the number of CPU cores used for computation, where $\text{n\_jobs}=-1$ utilizes all available cores.}=300, \\ 
& & contamination=0.001, n\_jobs=-1 \\
\hline
ADC2 & For positive peak: height=-5, & $\text{n\_estimators}$=300, \\
& prominence=5, distance=20 & contamination=0.00001, n\_jobs=-1 \\
& For negative peak: height=15, & \\ 
& prominence=15, distance=20 & \\
\hline
ADC3 & For positive peak: height=5, & $\text{n\_estimators}$=300, \\
& prominence=2, distance=100 & contamination=0.001, n\_jobs=-1 \\
& For negative peak: height=500, & \\ 
& prominence=500, distance=100 & \\
\hline
\end{tabular} \vspace{-1mm}
}
\end{table}

\begin{table} [t!]
\caption{Performance of each method.}
\label{res_met} 
\centering
{
\setlength\tabcolsep{1.0pt}
\begin{tabular}{|c|c|c|c|c|}
\hline
Signal & Total & Anomaly detection & Processing rate & Anomaly rate \\
& anomalies & time (in seconds) & (in samples/second) & (in \%) \\
\hline
\multicolumn{5}{|c|}{Signal Processing} \\
\hline
ADC1 & 349 & 0.1926 & 19,593,631.54 & 0.009247 \\
\hline
ADC2 & 14 & 0.1355 & 27,854,655.24 & 0.000371 \\
\hline
ADC3 & 1 & 0.0546 & 69,085,072.70 & 0.000026 \\
\hline
\multicolumn{5}{|c|}{Isolation Forest} \\
\hline
ADC1 & 3579 & 14.5215 & 259,890.61 & 0.0948 \\
\hline
ADC2 & 29 & 31.1416 & 121,188.64 & 0.0008 \\
\hline
ADC3 & 11 & 31.7886 & 118,722.27 & 0.0003 \\
\hline
\end{tabular} \vspace{-3mm}
}
\end{table}

\subsubsection{Signal Processing Analysis} 
\label{decision_sig_pro}
\begin{figure} [t!]
    \centering
    \includegraphics[width=\columnwidth, height=3.0cm]{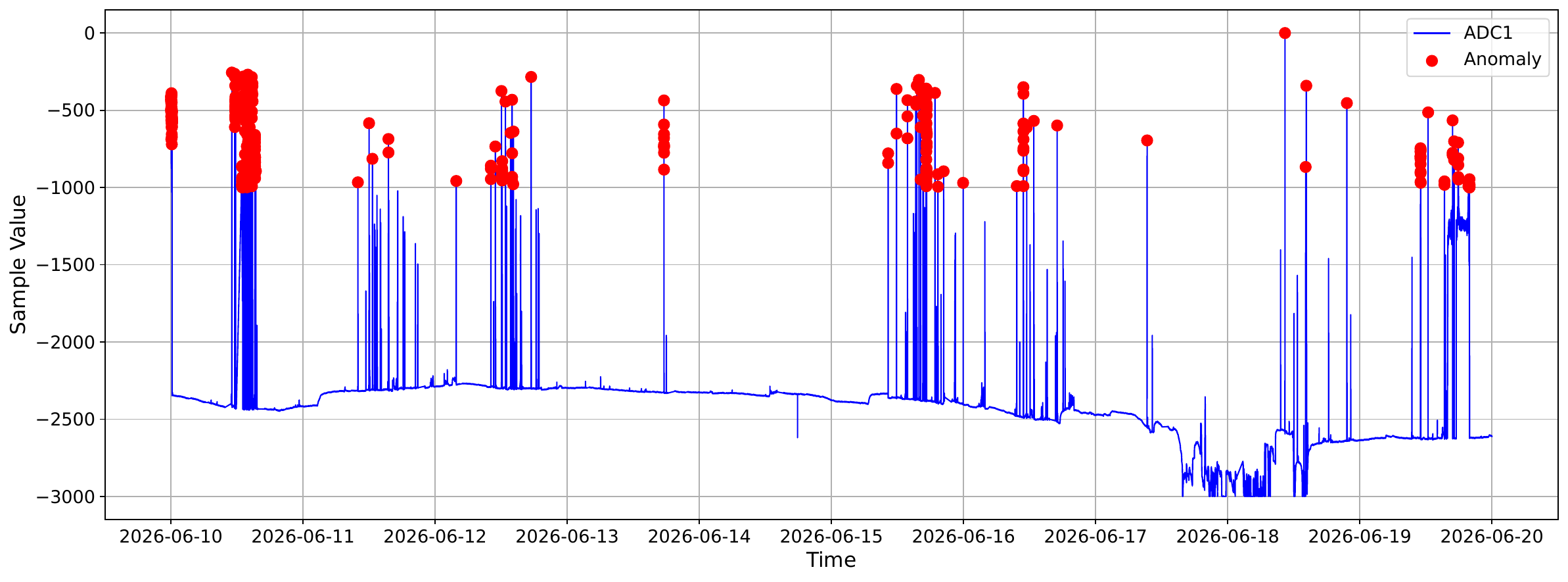}
    \caption{ADC1 anomalies detected using signal processing.}
    \label{adc1_anomaly_sp} \vspace{-3mm}
\end{figure}

\begin{figure} [t!]
    \centering
    \includegraphics[width=\columnwidth, height=3.0cm]{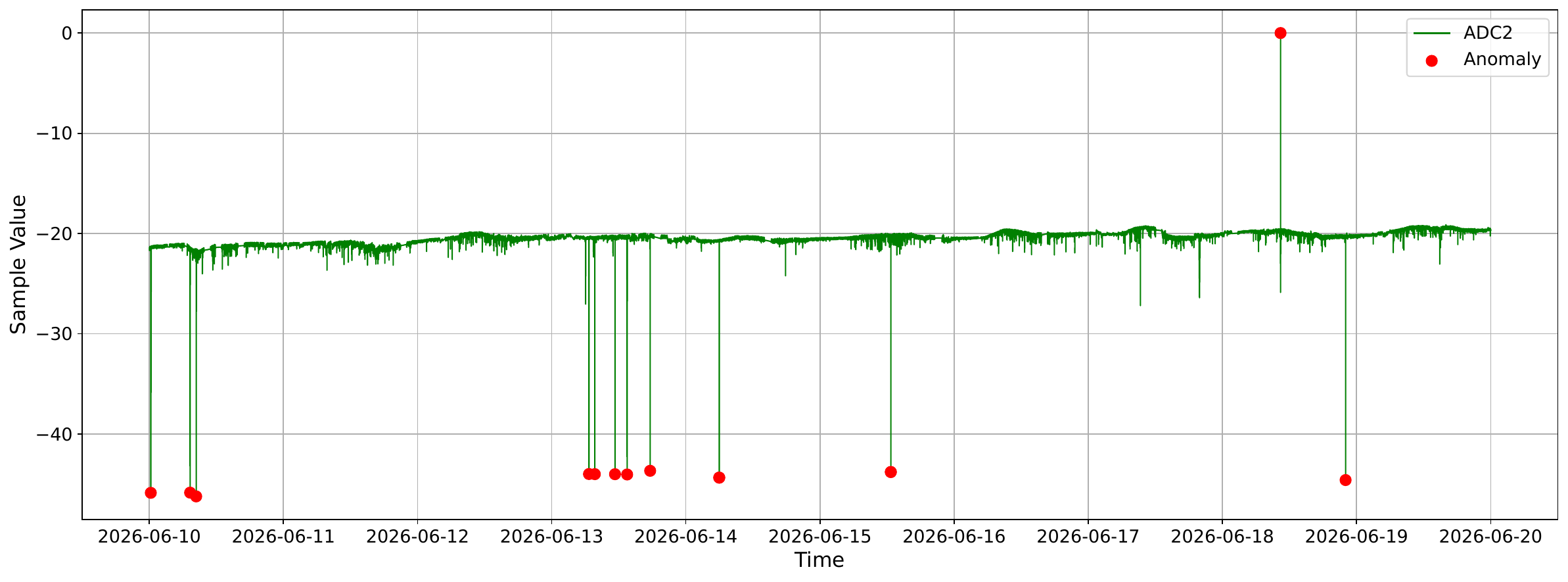}
    \caption{ADC2 anomalies detected using signal processing.}
    \label{adc2_anomaly_sp} \vspace{-3mm}
\end{figure}

\begin{figure} [t!]
    \centering
    \includegraphics[width=\columnwidth, height=3.0cm]{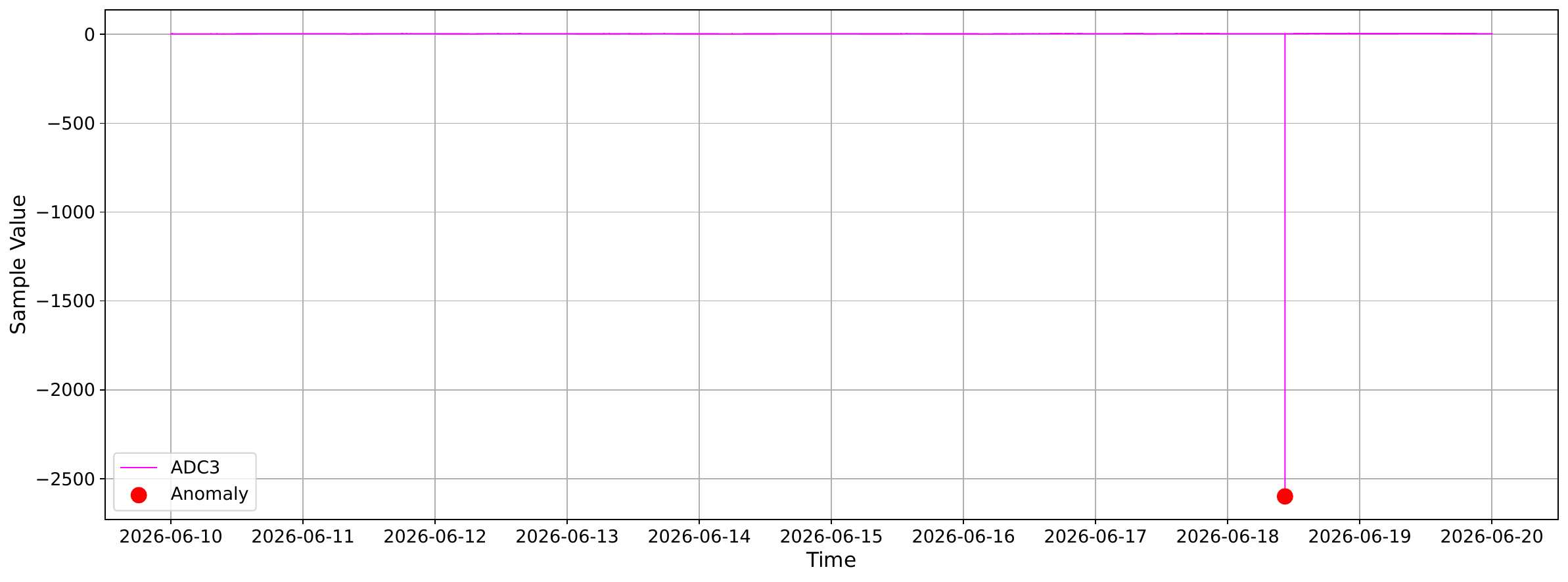}
    \caption{ADC3 anomalies detected using signal processing.}
    \label{adc3_anomaly_sp} \vspace{-3mm}
\end{figure}

As presented in Table~\ref{res_met}, the signal processing method detects 349, 14, and 1 anomalies in ADC1, ADC2, and ADC3, respectively. The anomaly detection time ranges from 0.0546 to 0.1926 seconds, with processing rates of 19.59-69.09 million samples/second. ADC3 achieves the shortest anomaly detection time and the highest processing rate, demonstrating the low computational cost of the method. The detected anomalies are illustrated in Fig.~\ref{adc1_anomaly_sp}, Fig.~\ref{adc2_anomaly_sp}, and Fig.~\ref{adc3_anomaly_sp}, respectively.

\subsubsection{Isolation Forest Analysis} 
\label{decision_if} 
\begin{figure} [t!]
    \centering
    \includegraphics[width=\columnwidth, height=3.0cm]{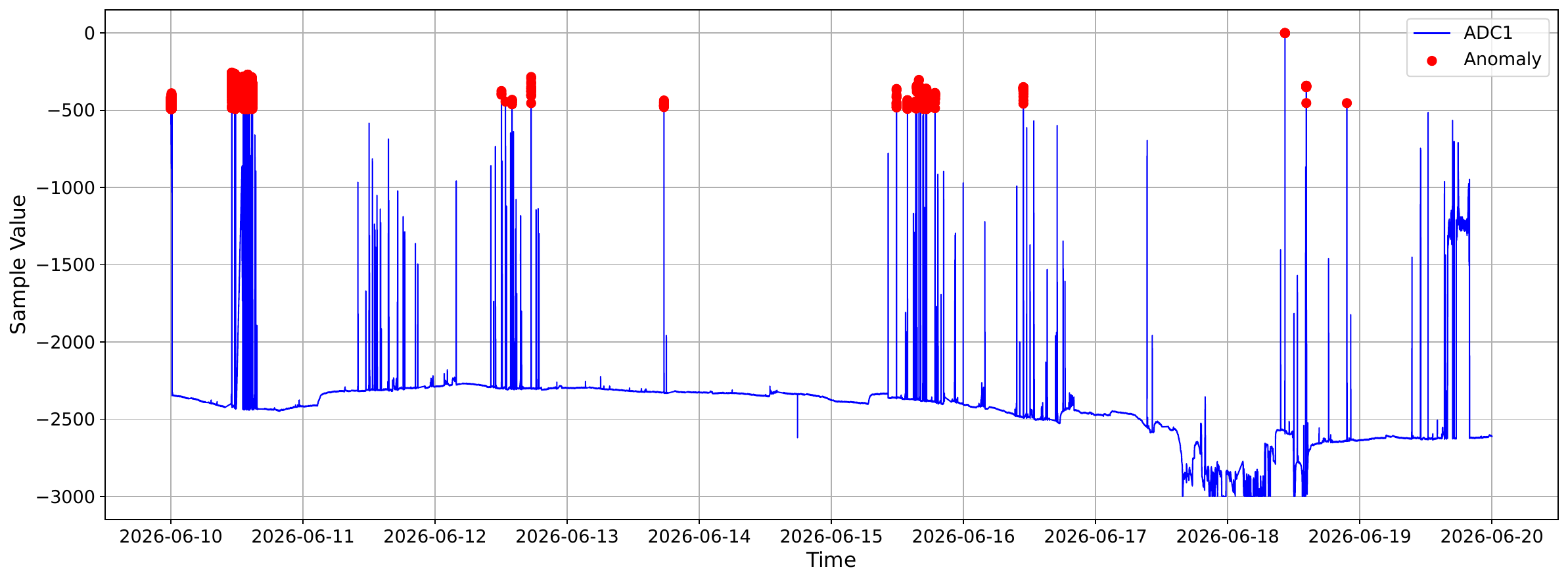}
    \caption{ADC1 anomalies detected using Isolation Forest.}
    \label{adc1_anomaly_if} \vspace{-3mm}
\end{figure}

\begin{figure} [t!]
    \centering
    \includegraphics[width=\columnwidth, height=3.0cm]{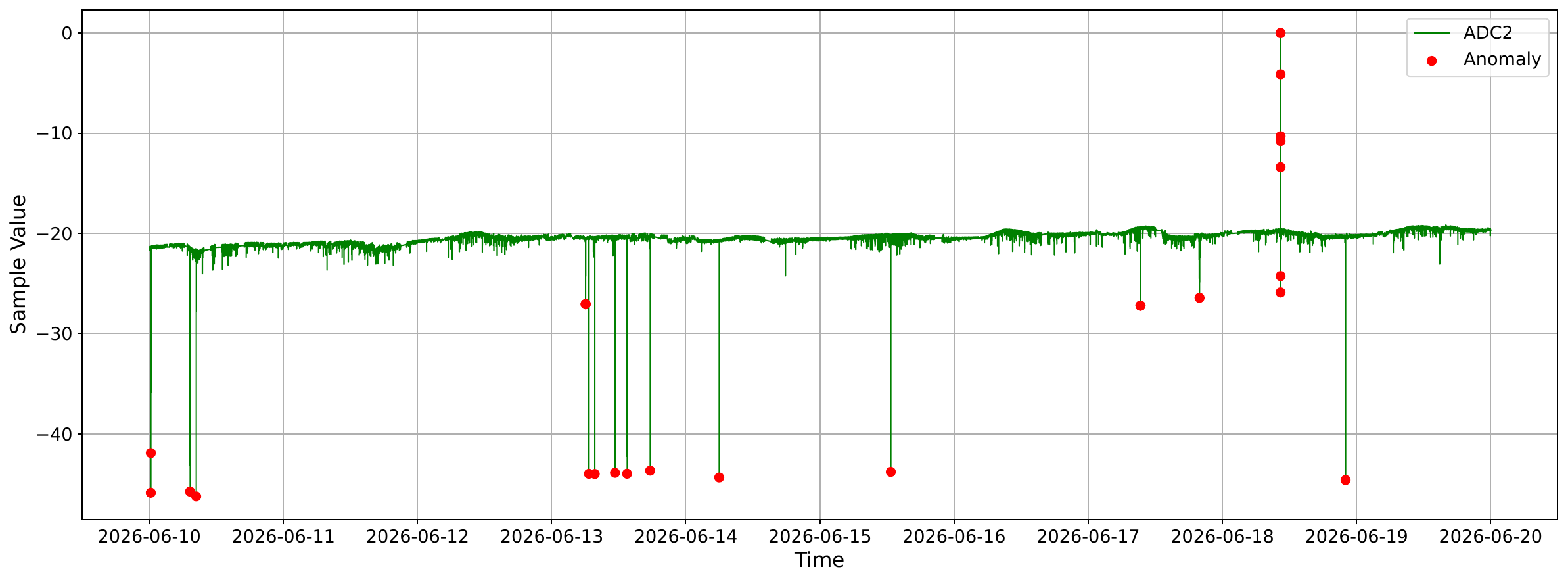}
    \caption{ADC2 anomalies detected using Isolation Forest.}
    \label{adc2_anomaly_if} \vspace{-3mm}
\end{figure}

\begin{figure} [t!]
    \centering
    \includegraphics[width=\columnwidth, height=3.0cm]{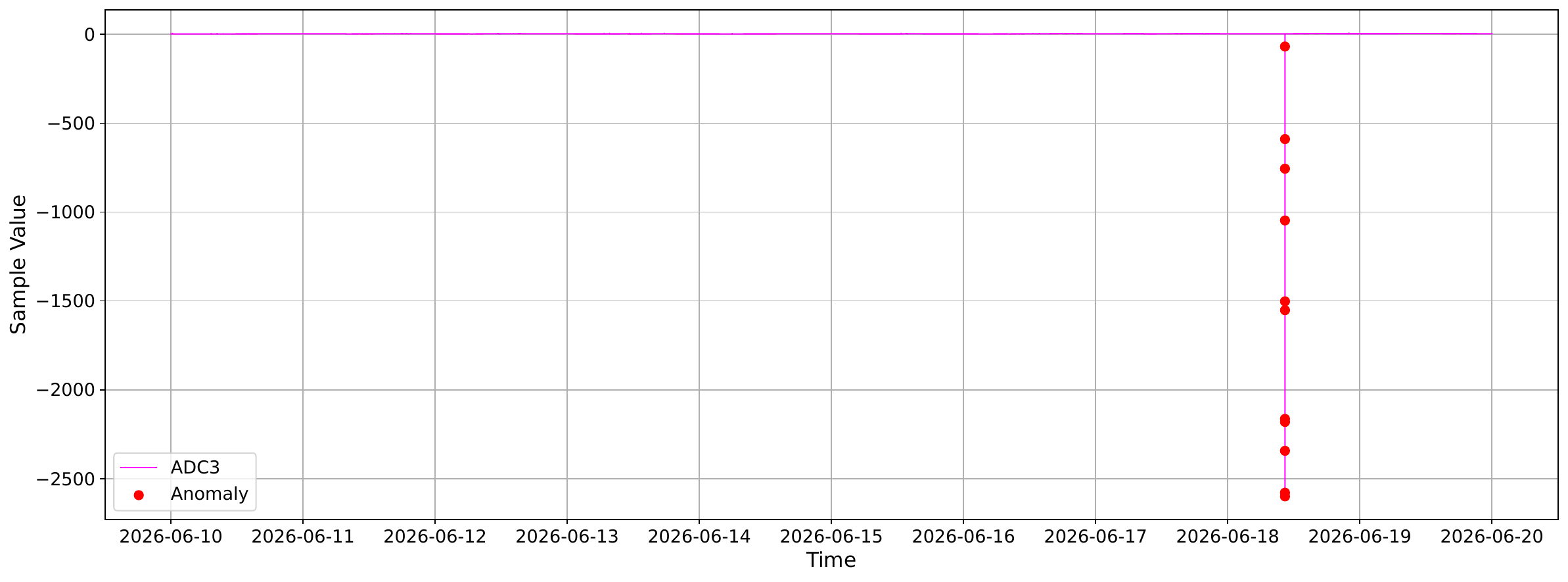}
    \caption{ADC3 anomalies detected using Isolation Forest.}
    \label{adc3_anomaly_if} 
\end{figure}

As presented in Table~\ref{res_met}, the Isolation Forest method identifies 3579, 29, and 11 anomalies in ADC1, ADC2, and ADC3, respectively. The anomaly detection time ranges from 14.52 to 31.79 seconds, while the processing rate varies between 118,722 and 259,891 samples/second. ADC1 requires the shortest anomaly detection time and achieves the highest processing rate, whereas ADC3 requires the longest anomaly detection time. Although computationally more demanding than the signal processing method, Isolation Forest provides a data-driven approach for identifying anomalous observations in the bridge sensor data. These results highlight the computational trade-off associated with the Isolation Forest approach. The detected anomalies are illustrated in Fig.~\ref{adc1_anomaly_if}, Fig.~\ref{adc2_anomaly_if}, and Fig.~\ref{adc3_anomaly_if}, respectively.

\subsection{Temporal Agreement Analysis} 
\label{temp_agree} 
This experiment evaluates how consistently signal processing and Isolation Forest methods identify anomalies at approximately the same time. Since the bridge sensor data do not contain ground-truth anomaly labels, the detections from SP and IF are also compared based on their temporal correspondence. An anomaly is considered temporally coincident when detections from the two methods occur within a specified sample-level tolerance. With a sampling rate of 5~Hz (0.2~seconds per sample), a primary tolerance of $\pm5$ samples ($\pm1$~second) is used. The analysis compares the number of matched anomalies and the corresponding agreement percentage between the two methods. Note that this analysis measures temporal agreement rather than anomaly-detection accuracy.

\begin{table}[t!]
\caption{Temporal agreement between each method using a $\pm5$ samples ($\pm1$ second) matching tolerance.}
\label{temporal_agre}
\centering
{
\setlength\tabcolsep{3.0pt}
\begin{tabular}{|l|c|c|c|c|c|}
\hline
Signal & SP & IF & Matched & SP agreement & IF agreement \\
& anomalies & anomalies & anomalies & (\%) & (\%) \\
\hline
ADC1 & 349 & 3579 & 100 & 28.65 & 2.79 \\
ADC2 & 14  & 29   & 2   & 14.29 & 6.90 \\
ADC3 & 1   & 11   & 1   & 100.00 & 9.09 \\
\hline
\end{tabular}\vspace{-3mm}
}
\end{table}

As presented in Table~\ref{temporal_agre}, 100 temporally coincident anomalies are identified between the two methods for ADC1, corresponding to 28.65\% of the signal processing detections and 2.79\% of the Isolation Forest detections. For ADC2, only two anomalies are temporally coincident, resulting in agreement ratios of 14.29\% and 6.90\%, respectively. For ADC3, the single anomaly detected by the signal processing method is also identified by Isolation Forest, resulting in 100\% agreement with respect to the signal processing detections. These results indicate that the two methods identify some temporally coincident anomalies, while a substantial number of detections remain method-specific.

\subsection{Controlled Anomaly-Injection Analysis} 
\label{cont_anom_inj} 
To further evaluate the sensitivity of the two anomaly detection methods, a controlled anomaly-injection experiment is conducted using the real bridge sensor data. To this end, three types of controlled perturbations, namely point spikes, short events, and level shifts, are injected into each ADC signal at moderate, strong, and extreme strengths. The point spikes represent abrupt changes at individual samples, short events represent temporary changes over a short sequence of samples, and level shifts represent a change in the signal level over a longer interval. Two anomalies are generated for each combination of anomaly type and strength, resulting in 18 injected anomalies per signal.

\begin{table}[t!]
\caption{Anomaly detection under controlled injections.}
\label{injec_res}
\centering
{
\setlength\tabcolsep{2.0pt}
\begin{tabular}{|l|c|c|c|c|c|}
\hline
Signal & Injected & SP detected & SP Recall (\%) & IF detected & IF Recall (\%) \\
\hline
ADC1 & 18 & 0 & 0.00 & 0 & 0.00 \\
ADC2 & 18 & 6 & 33.33 & 18 & 100.00 \\
ADC3 & 18 & 1 & 5.56 & 13 & 72.22 \\
\hline
\end{tabular}\vspace{-3mm}
}
\end{table}

As presented in Table~\ref{injec_res}, the signal processing method detects 0, 6, and 1 of the 18 injected anomalies in ADC1, ADC2, and ADC3, respectively, corresponding to anomaly-level recall values of 0.00\%, 33.33\%, and 5.56\%. Isolation Forest detects 0, 18, and 13 injected anomalies, resulting in recall values of 0.00\%, 100.00\%, and 72.22\%, respectively. These results demonstrate that the sensitivity of both methods varies across the different sensor signals. 

Note that the injected perturbations are synthetic and may not represent actual bridge anomalies; therefore, these results provide a controlled sensitivity analysis rather than a substitute for evaluation on real bridge conditions.

\section{Conclusions and Future Work} 
\label{cons}
This paper investigates anomaly detection in real-world bridge sensor data using signal processing and Isolation Forest. The results demonstrate that the two approaches provide complementary capabilities, with signal processing offering a simple and computationally efficient detection mechanism, while Isolation Forest provides a data-driven approach for identifying unusual sensor observations based on patterns in the bridge sensor data. Moreover, the temporal agreement analysis shows that both methods identify some anomalies at approximately the same time. The controlled anomaly-injection experiment demonstrates varying sensitivity across the sensor signals. Future work will focus on evaluating the proposed framework using data from multiple bridges and additional sensor modalities, as well as developing adaptive real-time detection approaches for continuous bridge monitoring.

\section*{Acknowledgement} 
The authors acknowledge Smart Sensor Systems AS for providing the data and supporting conference dissemination.

\bibliographystyle{IEEEtran}
\bibliography{references}

\end{document}